\documentclass[conference]{IEEEtran}

\IEEEoverridecommandlockouts

\pdfoutput=1

\usepackage{cite}
\usepackage{amsmath,amssymb,amsfonts}
\usepackage{algorithmic}
\usepackage{graphicx}
\usepackage{textcomp}
\def\BibTeX{{\rm B\kern-.05em{\sc i\kern-.025em b}\kern-.08em
    T\kern-.1667em\lower.7ex\hbox{E}\kern-.125emX}}
    
\usepackage[svgnames]{xcolor}
\usepackage{soul}

\begin{document}
%Online Evolving Granular Classifiers to Anomaly Behavior Detection of Log-based Systems for Predictive Maintenance in Data Centers}

\title{Comparison of Evolving Granular Classifiers applied to Anomaly Detection for Predictive Maintenance in Computing Centers}

\author{
\IEEEauthorblockN{Leticia Decker}
\IEEEauthorblockA{\textrm{University of Bologna}\\
INFN  \\
Bologna, Italy \\
leticia.deckerde@unibo.it}
\and
\IEEEauthorblockN{Daniel Leite}
\IEEEauthorblockA{\textrm{Federal University of Lavras}\\
Lavras, Brazil \\
daniel.leite@ufla.br}
\and
\IEEEauthorblockN{Fabio Viola}
\IEEEauthorblockA{\textit{INFN-CNAF}\\
Bologna, Italy \\
fabio.viola@cnaf.infn.it}
\and
\IEEEauthorblockN{Daniele Bonacorsi}
\IEEEauthorblockA{\textit{University of Bologna}\\
INFN\\
Bologna, Italy \\
daniele.bonacorsi@unibo.it}
}

\maketitle

\begin{abstract}
Log-based predictive maintenance of computing centers is a main concern regarding the worldwide computing grid that supports the CERN (European Organization for Nuclear Research) physics experiments. A \textit{log}, as event-oriented ad-hoc information, is quite often given as unstructured big data. Log data processing is a time-consuming computational task. The goal is to grab essential information from a continuously changeable grid environment to construct a classification model. Evolving granular classifiers are suited to learn from time-varying log streams and, therefore, perform online classification of the severity of anomalies. We formulated a 4-class online anomaly classification problem, and employed time windows between landmarks and two granular computing methods, namely, Fuzzy-set-Based evolving Modeling (FBeM) and evolving Granular Neural Network (eGNN), to model and monitor logging activity rate. The results of classification are of utmost importance for predictive maintenance because priority can be given to specific time intervals in which the classifier indicates the existence of high or medium severity anomalies. 
\end{abstract}


\section{Introduction}

Cloud Computing is a dynamically scalable computing paradigm that provides virtual resources through the internet to users aiming at saving costs while maintaining high availability of the resources. %\cite{Escalante}. 
Maintenance systems are usually based on offline statistical analysis of log records -- a preventive maintenance approach generally based on constant time intervals. Recently, online computational-intelligence-based frameworks, namely evolving fuzzy and neuro-fuzzy frameworks \cite{Skrjanc1,Cordovil,Garcia,Hyde,SilvaP}, combined with incremental machine-learning algorithms have been considered for on-demand anomaly detection, autonomous data classification, and predictive maintenance of an array of systems \cite{Venkatesan,Souza,Bezerra,Edw2020}.

The background of the present study is the main Italian WLCG (Worldwide Large-hadron-collider Computing Grid) data center, the Tier-1 hosted in Bologna. The WLCG involves 170 computing centers in 42 countries. It consists in a grid structure that supports the CERN physics experiments performed at the biggest particle accelerator in the world. The WLCG is organised in a multi-layer-distributed computing system, namely, tiers, which are numbered from 0 (at CERN facilities), to 3, in descending order of importance.

The Tier-1 Bologna is hosted by the Computing Center of the Italian Institute of Nuclear Physics (INFN-CNAF). Its infrastructure consists of approximately 40,000 CPU cores, 40 PB of disk storage, and 90 PB of tape storage, connected to the Italian (GARR) and European (GÉANT) research networks. Currently, it collects log data from 1197 machines in total. The computing center undergoes a technology exchange process aiming to provide all resources through cloud services. %\cite{INFN}. 

An important task of the WLCG computing center concerns the efficient use of the system resources, such as the resources used for system maintenance. In addition, user logs are noticed as service-oriented unstructured data. Large volumes of data are produced by a number of system logs, which makes the implementation of a general-purpose log-based predictive maintenance solution challenging \cite{chep}. Logging activity means the rate of lines written in a log file. The logging activity rate depicts the overall computing center behaviour in terms of service utilization. The rate may be used to monitor and detect service behavior anomalies to assist predictive maintenance focused on specific time intervals.

Being log data processing a highly time and resource-consuming application, an important issue concerns the identification of the most promising pieces of log files using an online machine learning strategy. This way, priority may be given to such log fragments, which maximizes the likelihood of finding useful information to the system maintenance.

Many attributes should be extracted from log timestamps, i.e., from time information related to the moment of log record writing. These attributes establish a sequence of encoded information that summarizes the system use in a time interval. The encoded information we emphasize are the mean, maximum difference between two consecutive means, variance, and minimum and maximum number of system accesses, as noticed in time windows between landmarks. We consider that the system activity impacts the log activity proportionally. Moreover, we propose a dynamic control-chart-based approach to label online data instances autonomously.

This paper addresses the dynamic log-based anomaly detection problem in cloud computing context as an unbalanced multiclass classification problem in which the classes refer to the severity of anomalies and the usual system behavior. A Fuzzy-Set-Based evolving Model (FBeM) \cite{Leite10} and an evolving Granular Neural Network (eGNN) \cite{Leite14} are developed from a stream of data dynamically extracted from time windows. FBeM and eGNN are supported by incremental learning algorithms able to keep their parameters and structure updated to reflect the current environment. As time windows are classified as containing anomalies, predictive maintenance can focus on that specific log information -- a small percentage of the whole. FBeM and eGNN are compared in classification accuracy, model compactness, and processing time.

The remainder of this paper is organised as follows. Section~\ref{sec:storm} presents a StoRM service use case. Section~\ref{sec:evolving} describes the evolving fuzzy and neuro-fuzzy approaches to anomaly classification. The methodology is given in Section~\ref{sec:meth}, and the results and discussions in Section~\ref{sec:er}. Conclusions and future research are outlined in Section~\ref{sec:cfw}.

\section{The Data Storage Service}
\label{sec:storm}
%\section{Related Literature}
%\label{sec:rl}

The predictive maintenance problem applied to computing centres is a CERN hot topic, highlighted by the programmed system upgrade, High-Luminosity Large Hadron Collider (HL-LHC) project. %\cite{CERN}.
This project intends to increase luminosity by a factor of 10 beyond the LHC’s initial design value. Once luminosity is proportional to the rate of particle collisions, increasing the luminosity will enlarge the volume of data and experiments substantially.  

Many efforts have been done to maintain the quality of service (QoS) of the WLCG. In particular, at the Tier-1 computing center at Bologna, supervised machine learning methods to predict anomalies of the StoRM service \cite{Giommi} have been considered by using ad-hoc data processing methods. Another initiative concerns a system based on the Elastic Stack Suite to collect, parse and catalogue log data, as well as classifying anomalies using an embedded unsupervised-learning tool \cite{Diotalevi}. With the goal of characterising log files, a clustering method based on Levenshtein distance was proposed in \cite{Rossi}, and a semi-supervised learning method in \cite{FUZZlog}. 

Close to the aim of the present paper, it is worth mentioning an offline prototype for anomaly detection using a One-Class method to classify anomalous time windows based on logging activity \cite{Minarini}. The present paper extends the offline anomalous time-window-based approach in \cite{Minarini} to evolving granular computing and online fuzzy and neuro-fuzzy frameworks. A Fuzzy-set-Based evolving Modeling, FBeM, approach \cite{Leite10} and an evolving Granular Neural Network, eGNN, approach \cite{Leite14} are compared considering online anomaly detection based on logging activity. Particularly, the log data stream in question comes from the StoRM Back-end file produced by the Tier-1 Bologna. FBeM and eGNN are applicable to and learn autonomously from any event-oriented log file. 

\begin{figure}[htp!]
    \begin{center}
       \includegraphics[width=1\columnwidth]{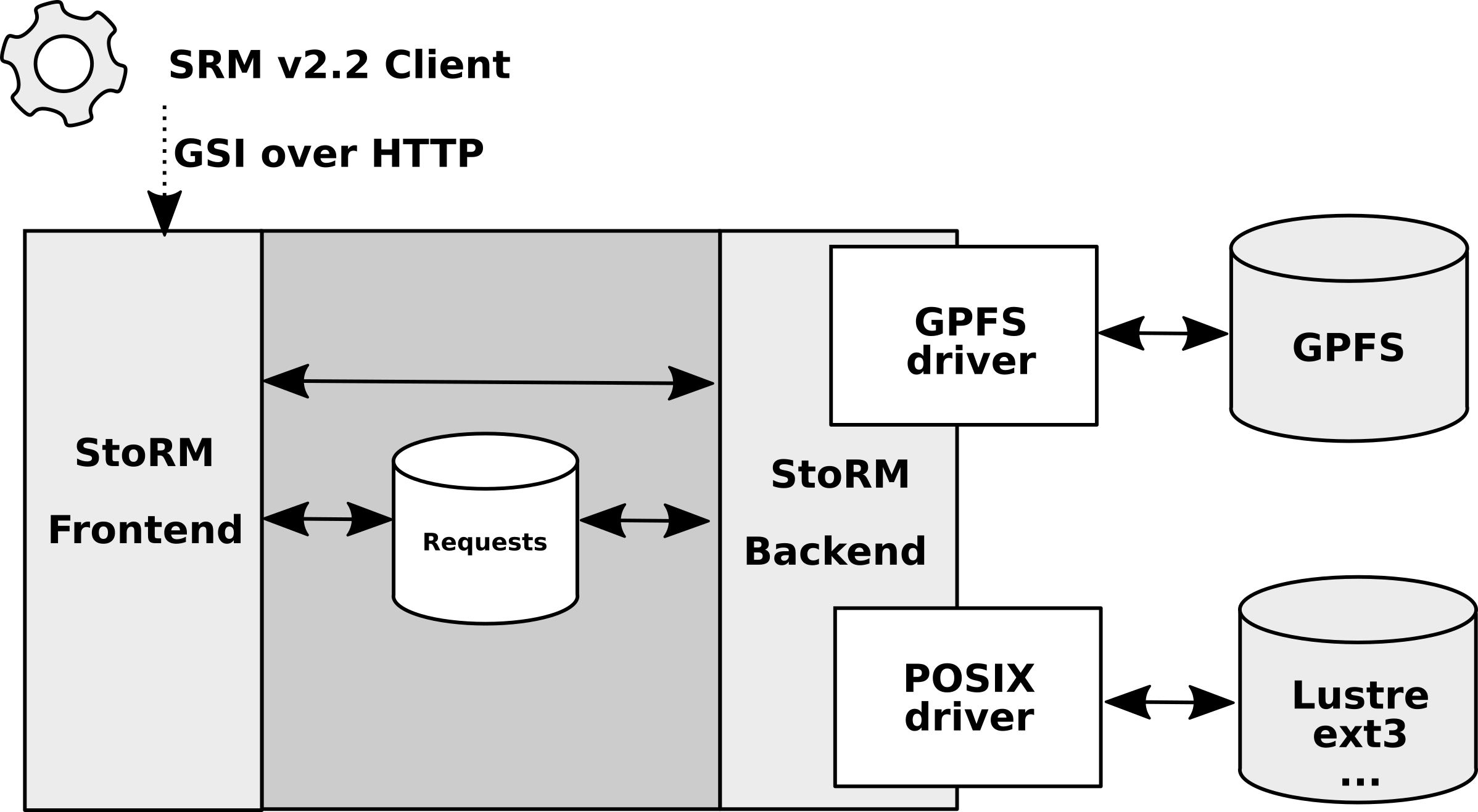}
    \end{center}
	\caption{SRM architecture and relations among its main modules: Front-end, Back-end, and databases}
	\label{fig1}
\end{figure}

Storage Resource Manager (SRM) is a service that provides the storage system used by INFN-CNAF. SRM aims to provide high performance to parallel file systems, such as the GPFS (the IBM General Parallel File System) and POSIX (Portable Operating System Interface), through a graphic interface to INFN-CNAF infrastructure users. SRM has modular architecture constitute by StoRM Front-end (FE), StoRM Back-end (BE), and databases (DB) \cite{Giommi}. The FE module manages user authentication, and store and load requests. The BE module is the main StoRM component regarding functionality. It executes all synchronous and asynchronous SRM operations, which allows interaction among grid elements. The SRM modules and their relations are shown in Fig.~\ref{fig1}.
\vspace{-4pt}
\vspace{-4pt}
All modules handle unstructured log data and two or more types of log writing mechanisms, such as scheduled and event-oriented mechanisms. There is, however, a common information to all log files which is the timestamp of the log record writing. Based on such timestamps, it is possible to group log files in time windows to monitor the logging activity rate. As scheduled log data follows a regular logging activity, we focus on event-oriented data.

%\clearpage

\section{Evolving Granular Classifiers}
\label{sec:evolving}

A diversity of evolving classifiers have been developed based on typicality and eccentricity concepts \cite{Soares2}, local strategies to smooth parameter changes \cite{Shaker}, self-organization of fuzzy models \cite{Gu}, ensembles of models \cite{Mirza}, scaffolding fuzzy type-2 models \cite{MPratama}, double-boundary granulation \cite{Charles}, semi-supervision concepts \cite{Kim}, interval analysis \cite{LeiteGra}, and uncertain data and multi-criteria optimization \cite{Cordovil,eogs}.

This paper compares FBeM and eGNN as evolving classifiers for anomaly detection. The classifiers: (i) evolve over time; (ii) are based on fuzzy rules, codified in a neural network or not; and (iii) granulate the data domain incrementally. The evolving classifiers adapt to new scenarios. In other words, their parameters and structure are learned on the fly, according to the data stream. Prototype initiation is needless. Hence, their accuracy tends to increase over time by means of recursive machine learning mechanisms \cite{Leite14}.

Data instances are classified using a set of fuzzy rules extracted from previous data. Each fuzzy rule is associated to a granule, and each class is represented by at least one granule and rule. The main difference between these methods is the strategy to generate and update granules and rules, which provide different decision boundaries.

Data granulation, granularity, and granular models come from the Granular Computing (GrC) theory \cite{GranularHandBook}. GrC is based on the fact that precision may be expensive and is usually an unnecessary task for modelling complex systems \cite{Leite14}. GrC provides a framework to express information in a higher level of abstraction. Granular learning, in this paper, means to fit new numerical data in granular models. Fuzz granules and aggregation neurons can be added, updated, removed and combined along the learning steps. Therefore, FBeM and eGNN capture new information from a data stream, and self-adapt to the current scenario.

% Each algorithm receives a stream $(x,y)^{[h]}$ of input vector $x$ and its classification $y$, used just to calculate the method accurancy. Both methods give as output a set of rules capable to classify a new input $x^{[h]}$ according to the anomaly level of the system behavior. In this work, the sistem behavior could be classified as normal, or in one of three anomaly levels shown in section \ref{sec:meth}

\subsection{FBeM: Fuzzy Set-Based Evolving Modeling}

FBeM is an online classifier based on fuzzy rules, which supports intense data streams. FBeM uses an incremental learning algorithm, and provides nonlinear class boundaries \cite{Leite14}.

Being an adaptive fuzzy classifier means that the model consists of a time-varying amount of elements (granules) of granularity $\rho$. In this paper, the $i$-th fuzzy set on the $j$-th domain is described by a trapezoidal membership function $A_{j}^{i} = (l_j^i, \lambda_j^i, \Lambda_j^i, L_j^i)$, which is shape-evolving within a maximum extension region, $E_{j}^{i}$. A datum $x_{j}$ may have partial membership on $A_{j}^{i}$. 

FBeM rules associate granules to class labels. The rule 

\vspace{-2pt}

\begin{eqnarray}
R^i(x):~\textrm{if}(x_1~ \textrm{is}~ A_1^i)~\textrm{and}\dots\textrm{and}~(x_n~\textrm{is}~ A_n^i)~\textrm{then}~(\Bar{y}~\textrm{is}~\hat{C}) \nonumber
\end{eqnarray}

\noindent holds within the limits of the granule. Time-varying rules $R^i$ compose the whole model $R = \{ R^1,\dots,R^i,\dots,R^c\}$. An input instance is denoted by $\textbf{x} = (x_1,\dots,x_j,\dots,x_n)$; and $\hat{C}$ is the estimated class. A time-indexed pair, $(\textbf{x},C)^{[h]}$, is an entry of the data stream -- being $C$ the actual class.

The most active rule for $(\textbf{x},y)^{[h]}$ comes from

\vspace{-2pt}
\begin{eqnarray}
\alpha^{\ast} = S(\alpha^{1},\dots, \alpha^{i}, \dots, \alpha^{c})
\end{eqnarray}{}
\vspace{-2pt}
\noindent with
\vspace{-2pt}
\begin{eqnarray}
\alpha^{i} = T(A_1^{i},\dots, A_j^{i}, \dots, A_n^{i})
\end{eqnarray}{}

\noindent in which $S$ and $T$ are the $min$ and $max$ operators.

The width of $A_{j}^{i}$ is

\vspace{-2pt}
\begin{eqnarray}
wdt(A_{j}^{i}) = L_{j}^{i}-l_{j}^{i}, & ~ \textrm{then} ~ & wdt(A_{j}^{i}) \leq \rho_{j},
\end{eqnarray}{}
\vspace{-2pt}

\noindent in which $\rho_{j}$ is the model granularity.

A different initial $\rho$ produces different classifiers. However, $\rho$ changes autonomously over time from

\vspace{-2pt}
\begin{eqnarray}
\rho^{[h]} = 
\begin{cases}{}
    (1+ \frac{r}{h_{r}})\rho^{[h-hr]}, &  \textit{$r > \eta$} \\
    (1- \frac{\eta - r}{h_{r}})\rho^{[h-hr]}, &  \textit{$r \leqslant \eta$}
\end{cases}
\label{updatarho}
\end{eqnarray}{}
\vspace{-2pt}

\noindent in which $\eta$ is the growth rate; $r = c^{[h]} - c^{[h-h_{r}]}$, and $c^{[h]}$ is the amount of granules at the $h$-th time step; $h > h_{r}$.

The fuzzy set $A_{j}^{i}$ can be expanded within the region $E_{j}^{i}$. The expansion region is defined by

\vspace{-2pt}
\begin{eqnarray}
E_{j}^{i}=[mp(A_{j}^{i})-\frac{\rho_{j}}{2}, mp(A_{j}^{i})+\frac{\rho_{j}}{2}],
\end{eqnarray}

\noindent in which 

\vspace{-2pt}
\begin{eqnarray}
mp(A_{j}^{i}) = \frac{\lambda_{j}^{i} +\Lambda_{j}^{i}}{2}
\end{eqnarray}

\noindent is the midpoint of $A_{j}^{i}$.

As $\rho$ changes over time, if at least one entry, $x_{j}$, of $\textbf{x}$ does not belong to some $E_{j}^{i}$, $\forall i$, then a new rule is created. Its membership functions are

\vspace{-2pt}
\begin{eqnarray}
A_{j}^{c+1} = (l_{j}^{c+1}, \lambda_{j}^{c+1}, \Lambda_{j}^{c+1}, L_{j}^{c+1}) = (x_{j}, x_{j}, x_{j}, x_{j}).
\end{eqnarray}{}
\vspace{-2pt}

\noindent Otherwise, if a new $\textbf{x}$ belongs to an $E^{i}$, then the $i$-th rule is updated according to one of the following options:

\vspace{-2pt}
\begin{eqnarray}
\begin{array}{ll}
\textrm{if} ~x^{[h]} \in [mp(A_{j}^{i})-\frac{\rho_{j}}{2}, l_{j}^{i}]) &\textrm{then}~ l_{j}^{i}(new) = x^{[h]}\\
\textrm{if}~ x^{[h]} \in [l_{j}^{i}, \lambda_{j}^{i}]
&\textrm{then}~ \lambda_{j}^{i}(new) = x^{[h]}\\
\textrm{if}~ x^{[h]} \in [\lambda_{j}^{i}, mp(A_{j}^{i})]
&\textrm{then}~ \lambda_{j}^{i}(new) = x^{[h]}\\
\textrm{if}~ x^{[h]} \in [mp(A_{j}^{i}), \Lambda_{j}^{i}]
&\textrm{then}~ \Lambda_{j}^{i}(new) = x^{[h]}\\
\textrm{if}~ x^{[h]} \in [\Lambda_{j}^{i}, L_{j}^{i}]
&\textrm{then}~ \Lambda_{j}^{i}(new) = x^{[h]}\\
\textrm{if}~ x^{[h]} \in [L_{j}^{i}, mp(A_{j}^{i})+\frac{\rho_{j}}{2}]
&\textrm{then}~ L_{j}^{i}(new) = x^{[h]}\\
\end{array}
\end{eqnarray}
\vspace{-2pt}
%\begin{algorithmic}
    %\STATE if $x^{[h]} \in [mp(A_{j}^{i})-\frac{\rho_{j}}{2}, l_{j}^{i}])$ ~~~~ then $l_{j}^{i}(new) = x^{[h]}$
    %\STATE if $x^{[h]} \in [l_{j}^{i}, \lambda_{j}^{i}]$ ~~~~~~~~~~~~~then $\lambda_{j}^{i}(new) = x^{[h]}$
    %\STATE if $x^{[h]} \in [\lambda_{j}^{i}, mp(A_{j}^{i})]$ ~~~~~~then $\lambda_{j}^{i}(new) = x^{[h]}$
    %\STATE if $x^{[h]} \in [mp(A_{j}^{i}), \Lambda_{j}^{i}]$~~~~~~ then $\Lambda_{j}^{i}(new) = x^{[h]}$
    %\STATE if $x^{[h]} \in [\Lambda_{j}^{i}, L_{j}^{i}]$ ~~~~~~~~~~~~~ then $\Lambda_{j}^{i}(new) = x^{[h]}$
    %\STATE if $x^{[h]} \in [L_{j}^{i}, mp(A_{j}^{i})+\frac{\rho_{j}}{2}]$ then $L_{j}^{i}(new) = x^{[h]}$
%\end{algorithmic}

For a complete description on updating, removing, conflict resolution, and merging procedures see \cite{Leite10}. The FBeM algorithm is summarized below. Notice that we introduced a weak label to instances by means of a control chart approach, as described in the methodology section.

\vspace{5pt}

\hrule
\vspace{6pt}
\textbf{FBeM Learning: Fuzzy Set-Based Evolving Modeling}
\vspace{3pt}
\hrule
\vspace{4pt}
\begin{algorithmic}[1]
    \STATE \textbf{set} $\rho^{[0]}$, $h_r$, $\eta$;
    \STATE \textbf{for} $h = 1 \dots$
    \STATE ~~\textbf{read} $\textbf{x}^{[h]}$;
    \STATE ~~\textbf{use} control chart to label $\textbf{x}^{[h]}$ with $C^{[h]}$;
    \STATE ~~\textbf{provide} estimated class $\hat{C}^{[h]}$;
    \STATE ~~\textbf{compute} estimation error $\epsilon^{[h]} = C^{[h]} - \hat{C}^{[h]}$;
    \STATE ~~\textbf{if} $\textbf{x}^{[h]} \notin E^i ~ \forall i$ OR $C^{[h]}$ is new \textbf{then}
    \STATE ~~~~\textbf{create} new granule $\gamma^{c+1}$, with class $C^{[h]}$;
    \STATE ~~\textbf{else}
    \STATE ~~~~\textbf{update} the most active granule $\gamma^{i*}$ whose class is $C^{[h]}$;
    \STATE ~~\textbf{end if}
    \STATE ~~\textbf{delete} $\textbf{x}^{[h]}$;
    \STATE ~~\textbf{if} $h=\beta h_r$, $\beta= 1, \dots$ \textbf{then}
    \STATE ~~~~\textbf{merge} similar granules if needed;
    \STATE ~~~~\textbf{update} granularity $\rho$;
    \STATE ~~~~\textbf{delete} inactive granules if needed;
    \STATE ~~\textbf{end if}
    \STATE \textbf{end for}
\end{algorithmic}
\vspace{3pt}
\hrule
\vspace{4pt}

\subsection{eGNN: Evolving Granular Classification Neural Network}

eGNN is a neuro-fuzzy granular network constructed incrementally from an online data stream \cite{Leite14}. Its processing units are fuzzy neurons and granules, used to encode a set of fuzzy rules extracted from the data, and the neural processing conforms with a fuzzy inference system. The network architecture results from a gradual construction. The consequent part of an eGNN rule is a class in this paper.

\subsubsection{Neural Network}

\begin{figure}[ht]
	\begin{center}
		{\includegraphics[scale=0.46]{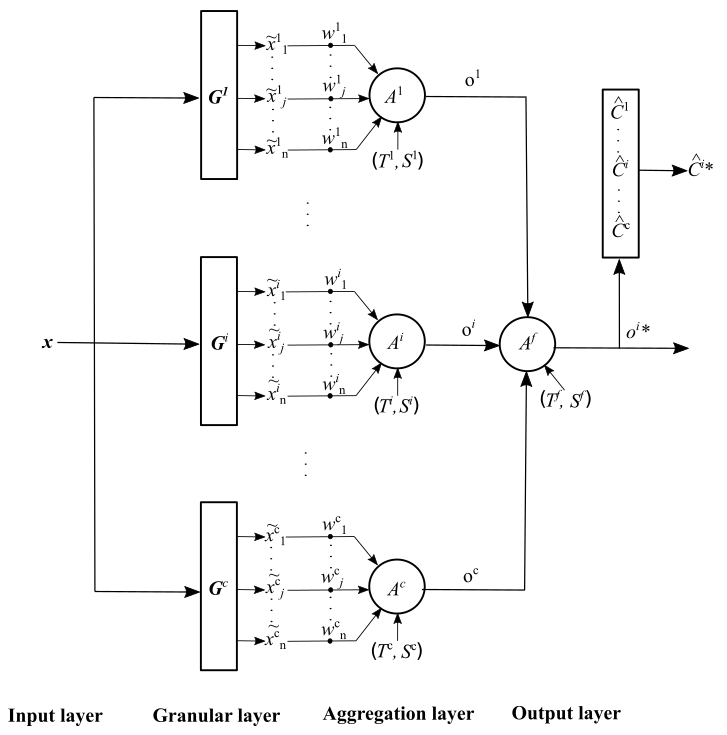}}
		\caption{eGNN: Evolving neuro-fuzzy network architecture for classification} \label{Fig7}
	\end{center}
\end{figure}

Consider that the data stream $(\textbf{x},y)^{[h]}$, $h = 1,...$, is measured from an unknown function $f$. Inputs $x_j$ and output $y$ are numerical data and a class. Figure \ref{Fig7} shows a four-layer eGNN model. The input layer receives $\textbf{x}^{[h]}$. The granular layer is a set of granules $G^i_j$, $i = 1, \dots, c$, stratified from input data, forming a fuzzy partition of the $j$-th input domain. A granule $G^i= G^i_1 \times \dots \times G^i_n$ is a fuzzy relation, i.e., a multidimensional fuzzy set in $X_1 \times \dots \times X_n$. Thus, $G^i$ has membership function $G^i(x) = min\{G_1^i(x_1),\dots,G_n^i(x_n)\}$ in $X_1 \times \dots \times X_n$. 

Similarity degrees $\widetilde{\textbf{x}}^{i} = (\widetilde{x}^{i}_1, \dots, \widetilde{x}^{i}_n)$ is the result of matching between $\textbf{x} = (x_1, \dots, x_n)$ and trapezoidal fuzzy sets $G^i = (G_1^i,\dots,G_n^i)$, with $G_j^i= (\underline{\underline{g}}_j^i,  \underline{g}_j^i, \overline{g}_j^i, \overline{\overline{g}}_j^i)$. In general, data and granules can be trapezoidal objects. A similarity measure to quantify the match between a numerical instance (the case of this paper) and the current knowledge is \cite{Leite14}:

\vspace{-2pt}

\begin{eqnarray}
\widetilde{x}_j^i = 1 - \frac{ |\underline{\underline{g}}_j^i \hspace{-1pt} - \hspace{-1pt} x_j| \hspace{-1pt} + \hspace{-1pt} |\underline{g}_j^i \hspace{-1pt} - \hspace{-1pt} x_j| \hspace{-1pt} + \hspace{-1pt} |\overline{g}_j^i \hspace{-1pt} - \hspace{-1pt} x_j| \hspace{-1pt} + \hspace{-1pt} |\overline{\overline{g}}_j^i \hspace{-1pt} - \hspace{-1pt} x_j| }{ 4 (\max(\overline{\overline{g}}_j^i, x_j) - \min(\underline{\underline{g}}_j^i, x_j)) } . \label{sdg}
\end{eqnarray}
\vspace{-2pt}

The aggregation layer is composed by neurons $A^i$. A fuzzy neuron $A^i$ combines weighted similarity degrees $(\widetilde{x}^{i}_1 w^{i}_1, \dots$, $\widetilde{x}^{i}_n w^{i}_n)$ into a single value $o^i$, which refers to the level of activation of $R^i$. The output layer processes $(o^1, \dots, o^c)$ using a neuron $A^f$ that performs the maximum S-norm. The class $C^{i*}$ of the most active rule $R^{i*}$ is the output.

Under assumption on specific weights and neurons, fuzzy rules extracted from eGNN are of the type

\vspace{-5pt}

\begin{eqnarray}
R^i(x):\textrm{if }(x_1~ \textrm{is}~ G_1^i)~\textrm{and}~\dots~\textrm{and}~(x_n~\textrm{is}~ G_n^i)~\textrm{then}~(\hat{y}~\textrm{is}~\hat{C}^i) \nonumber
\end{eqnarray}

\subsubsection{Neuron Model}

\begin{figure}[ht]
	\begin{center}
		{\includegraphics[scale=0.22]{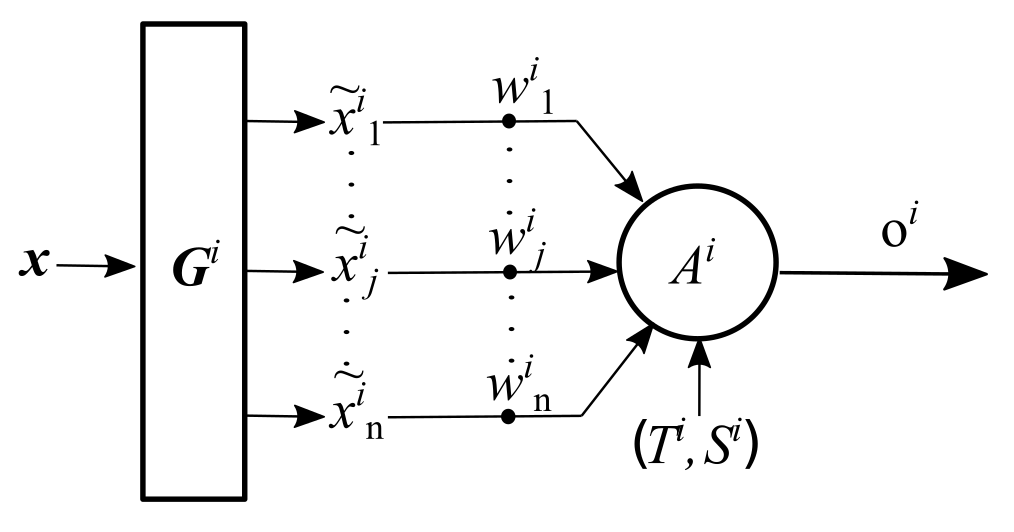}}
		\caption{Fuzzy neuron model} \label{Fig8}
	\end{center}
\end{figure}

Fuzzy neurons are neuron models based on aggregation operators. Aggregation operators $A:[0,1]^n\rightarrow[0,1]$, $n > 1$, combine input values in the unit hyper-cube $[0,1]^n$ into a single value in $[0,1]$. They must satisfy the following: monotonicity in all arguments and boundary conditions \cite{Leite14}. This study uses the minimum and maximum operators only \cite{Pedrycz,Beliakov}. Figure \ref{Fig8} shows an example of fuzzy neuron in which synaptic processing is given by the T-norm product, and the aggregation operator $A^i$ is used to combine individual inputs. The output $o^i$ is $A^i(\widetilde{x}^i_1 w^i_1, \dots, \widetilde{x}^i_n w^i_n)$.

%\begin{eqnarray}
%o = A(\widetilde{x}_1 w_1, \dots, \widetilde{x}_n w_n).
%\end{eqnarray}

%Figure~\ref{Fig8} shows the fuzzy neuron model. It produces a diversity of nonlinear mapping functions between input and output, depending on the chosen weights $w$ and aggregation operator $A^i$.
%Fig8.jpg
\vspace{-4pt}
\vspace{-4pt}

\subsubsection{Granular Region}

As $G^i_j$ is a trapezoidal membership function, its support, core, midpoint, and width are 

\begin{eqnarray}
\textrm{supp}(G^i_j) = [\underline{\underline{g}}^i_j,\overline{\overline{g}}^i_j], \label{supp} ~~~~
\textrm{core}(G^i_j) = [\underline{g}^i_j,\overline{g}^i_j] \\
\textrm{mp}(G^i_j) = \frac{\underline{g}^i_j+\overline{g}^i_j}{2}, ~~~~
\textrm{wdt}(G^i_j) = \overline{\overline{g}}^i_j - \underline{\underline{g}}^i_j.
\end{eqnarray}
\vspace{-2pt}

It is possible to expand the width of fuzzy sets $G^i_j$ within the area $E^i_j$ delimited by $\rho$, i.e., wdt$(G^i_j) \leq \rho$. $E^i_j$ is given by $[\textrm{mp}(G^i_j) - \frac{\rho}{2}, \textrm{mp}(G^i_j) + \frac{\rho}{2}]$. Clearly, wdt$(G^i_j) \leq$ wdt$(E^i_j)$. 

%\begin{eqnarray}
%E^i_j &=& [\textrm{mp}(G^i_j) - \frac{\rho}{2}, \textrm{mp}(G^i_j) + \frac{\rho}{2}]. \label{er}
%\end{eqnarray}

%\noindent It follows that wdt$(G^i_j) \leq$ wdt$(E^i_j)$ $\forall j,i$. 
%Values of $\rho$ allow different representations of the same problem at different levels of detail.

\subsubsection{$\rho$ Update}

The value of $\rho$ affects the information granularity, and consequently the model accuracy. $\rho \in [0,1]$ is used to control the size of expansion regions. 

eGNN starts with an empty rule base, and $\rho^{[0]} = 0.5$ is used as default. Let $r$ be the number of rules created in $h_r$ steps, and $\eta$ be a reference rate. If the number of rules grows faster than the rate $\eta$, then $\rho$ is increased, otherwise $\rho$ is reduced, as in Eq. \eqref{updatarho}. Appropriate values for $\rho$ are found autonomously. If $\rho = 1$, then eGNN is structurally stable, but unable to capture abrupt changes. Conversely, if $\rho = 0$, then eGNN overfits the data causing excessive model complexity. Adaptability is reached from intermediate values.

Reducing $\rho$ requires reduction of big granules according to

\vspace{-2pt}
\begin{eqnarray}
\begin{array}{ll}
\textrm{if} ~\textrm{mp}(G^i_j) \hspace{-1pt} - \hspace{-1pt} \frac{\rho \textrm{(new)}}{2} \hspace{-1pt} > \hspace{-1pt} \underline{\underline{g}}^i_j &\textrm{then}~ \underline{\underline{g}}^i_j \textrm{(new)} \hspace{-1pt} = \hspace{-1pt} \textrm{mp}(G^i_j) \hspace{-1pt} - \hspace{-1pt} \frac{\rho \textrm{(new)}}{2}\nonumber\\
\textrm{if}~ \textrm{mp}(G^i_j) \hspace{-1pt} + \hspace{-1pt} \frac{\rho \textrm{(new)}}{2} \hspace{-1pt} < \hspace{-1pt} \overline{\overline{g}}^i_j\nonumber &\textrm{then}~ \overline{\overline{g}}^i_j \textrm{(new)} \hspace{-1pt} = \hspace{-1pt} \textrm{mp}(G^i_j) \hspace{-1pt} + \hspace{-1pt} \frac{\rho \textrm{(new)}}{2}
\end{array}
\end{eqnarray}

\noindent Cores $[\underline{g}_j^i,\overline{g}_j^i]$ are handled in a similar way.

\subsubsection{Developing Granules}

If the support of at least one entry of $\textbf{x}$ is not enclosed by expansion regions $(E^i_1, \dots, E^i_n)$, eGNN generates a new granule, $G^{c+1}$. This new granule is constituted by fuzzy sets whose parameters are

\vspace{-2pt}

\begin{eqnarray}
G^{c+1}_j = (\underline{\underline{g}}_j^{c+1}, \underline{g}_j^{c+1}, \overline{g}_j^{c+1}, \overline{\overline{g}}_j^{c+1}) = (x_j, x_j, x_j, x_j).
\end{eqnarray}

\vspace{-2pt}

Updating granules consists in expanding or contracting the support and the core of fuzzy sets $G^i_j$. In particular, $G^i$ is chosen from $arg~max (o^1,\dots, o^c)$.

Adaptation proceeds depending on where $x_j$ in placed in relation to $G^i_j$.

\vspace{-2pt}

\begin{eqnarray}
\begin{array}{llllllll} \hspace{-1pt} \textrm{if } \hspace{-3pt} & x_j \in [\textrm{mp}(G^i_j)-\frac{\rho}{2},\underline{\underline{g}}^i_j] & \hspace{-4pt} ~ \textrm{then } ~~ \underline{\underline{g}}^i_j \textrm{(new)} = x_j \nonumber \\
\hspace{-1pt} \textrm{if } \hspace{-3pt} & x_j \in [\textrm{mp}(G^i_j)-\frac{\rho}{2},\textrm{mp}(G^i_j)] & \hspace{-4pt} ~ \textrm{then } ~~ \underline{g}^i_j \textrm{(new)} = x_j \nonumber \\
\hspace{-1pt} \textrm{if } \hspace{-3pt} & x_j \in [\textrm{mp}(G^i_j),\textrm{mp}(G^i_j)+\frac{\rho}{2}] & \hspace{-4pt} ~ \textrm{then } ~~ \underline{g}^i_j \textrm{(new)} = \textrm{mp}(G^i_j) \nonumber \\
\hspace{-1pt} \textrm{if } \hspace{-3pt} & x_j \in [\textrm{mp}(G^i_j)-\frac{\rho}{2},\textrm{mp}(G^i_j)] & \hspace{-4pt} ~ \textrm{then } ~~ \overline{g}^i_j \textrm{(new)} = \textrm{mp}(G^i_j) \nonumber \\
\hspace{-1pt} \textrm{if } \hspace{-3pt} & x_j \in [\textrm{mp}(G^i_j),\textrm{mp}(G^i_j)+\frac{\rho}{2}] & \hspace{-4pt} ~ \textrm{then } ~~ \overline{g}^i_j \textrm{(new)} = x_j \nonumber \\
\hspace{-1pt} \textrm{if } \hspace{-3pt} & x_j \in [\overline{\overline{g}}^i_j,\textrm{mp}(G^i_j)+\frac{\rho}{2}] & \hspace{-4pt} ~ \textrm{then } ~~ \overline{\overline{g}}^i_j \textrm{(new)} = x_j \nonumber
\end{array}
\end{eqnarray}

\noindent Operations on core parameters, $\underline{g}^i_j$ and $\overline{g}^i_j$, require additional adaptation of the midpoint

\begin{eqnarray}
\textrm{mp}(G^i_j)\textrm{(new)} = \frac{\underline{g}^i_j \textrm{(new)} + \overline{g}^i_j \textrm{(new)}}{2}.
\end{eqnarray}

\vspace{-2pt}

Therefore, support contractions may be needed:

\begin{eqnarray}
\begin{array}{ll}
\hspace{-4pt} \textrm{if} \hspace{3pt} \textrm{mp}(G^i_j)\textrm{(new)} \hspace{-1pt} - \hspace{-1pt} \frac{\rho}{2} \hspace{-1pt} > \hspace{-1pt} \underline{\underline{g}}^i_j \hspace{3pt} \textrm{then} \hspace{3pt} \underline{\underline{g}}^i_j \textrm{(new)} \hspace{-1pt} = \hspace{-1pt} \textrm{mp}(G^i_j)\textrm{(new)} \hspace{-1pt} - \hspace{-1pt} \frac{\rho}{2} \nonumber \\
\hspace{-4pt} \textrm{if} \hspace{3pt} \textrm{mp}(G^i_j)\textrm{(new)} \hspace{-1pt} + \hspace{-1pt} \frac{\rho}{2} \hspace{-1pt} < \hspace{-1pt} \overline{\overline{g}}^i_j \hspace{3pt} \textrm{then} \hspace{3pt} \overline{\overline{g}}^i_j \textrm{(new)} \hspace{-1pt} = \hspace{-1pt} \textrm{mp}(G^i_j)\textrm{(new)} \hspace{-1pt} + \hspace{-1pt} \frac{\rho}{2}. \nonumber
\end{array}
\end{eqnarray}
\vspace{-2pt}

\subsubsection{Updating Neural Network Weights}

$w^i_j \in [0,1]$ is proportional to the importance of the $j$-th attribute of $G_j^i$ to the neural network output. 
%If $w_j^i=1$, then the output is not affected. A relatively lower value of $w_j^i$ discounts the impact of the respective attribute. The procedure described below assigns lower weight values to less helpful attributes.
When a new granule $G^{c+1}$ is generated, weights are set as $w^{c+1}_j = 1$, $\forall j$. 
%If it is known \emph{a priori} that different attributes have different importance, then values for $w^{c+1}_j$ can be chosen in a way to reflect that.

The updated $w^i_j$, associated to the most active granule $G^i$, $i = arg~max (o^1, \dots, o^c)$, are 
\vspace{-2pt}
\begin{eqnarray}
w^i_j(\textrm{new}) = w^i_j(\textrm{old}) - \beta^i \widetilde{x}^{i}_j |\epsilon|. \label{wei}
\end{eqnarray}
\vspace{-2pt}

\noindent in which $\widetilde{x}^{i}_j$ is the similarity to $G^{i}_j$; $\beta^i$ depends on the number of right ($Right^i$) and wrong ($Wrong^i$) classifications

\vspace{-2pt}
\begin{eqnarray}
\beta^i = \frac{Wrong^i}{Right^i+Wrong^i} & ~ and ~ & \epsilon^{[h]} = C^{[h]} - \hat{C}^{[h]} \nonumber
\end{eqnarray}{}
\vspace{-2pt}

\noindent in which $\epsilon^{[h]}$ is the current estimation error, and $C^{[h]}$ is a weak label provided by the control chart approach described in the methodology section.

\subsubsection{Learning Algorithm}

The learning algorithm to evolve eGNN classifiers is given below.

~~

\hrule
\vspace{6pt}
\textbf{eGNN Learning: Evolving Granular Neural Network}
\vspace{3pt}
\hrule
\vspace{4pt}
\begin{algorithmic}[1]
\STATE \textbf{select} a type of neuron for the aggregation layer
\STATE \textbf{set} parameters $\rho^{[0]}$, $h_r$, $\eta$;
\STATE \textbf{read} instance $\textbf{x}^{[h]}$, $h = 1$;
\STATE \textbf{use} control chart to label $\textbf{x}^{[h]}$ with $C^{[h]}$;
\STATE \textbf{create} granule $G^{c+1}$, neurons $A^{c+1}$, $A^{f}$, and connections;
\STATE \textbf{for} $h = 2, \dots$ \textbf{do}
\STATE ~~~~\textbf{read} and  \textbf{feed-forward} $\textbf{x}^{[h]}$ through the network;
\STATE ~~~~\textbf{compute} rule activation levels $(o^1, \dots, o^c)$;
\STATE ~~~~\textbf{aggregate} activation using $A^f$ to get estimate $\hat{C}^{[h]}$;
\STATE ~~~~ // the class $C^{[h]}$ becomes available;
\STATE ~~~~\textbf{compute} output error $\epsilon^{[h]} = C^{[h]} - \hat{C}^{[h]}$;
\STATE ~~~~\textbf{if} $\textbf{x}^{[h]}$ \textbf{is not} $E^i ~ \forall i$ \textbf{or} $\epsilon^{[h]} \neq 0$ \textbf{then}
\STATE ~~~~~~~~\textbf{create} granule $G^{c+1}$, neuron $A^{c+1}$, connections;
\STATE ~~~~~~~~\textbf{associate} $G^{c+1}$ to $C^{[h]}$;
\STATE ~~~~\textbf{else}
\STATE ~~~~~~~~\textbf{update} $G^{i*}$, $i^* = arg~max (o^1, \dots, o^c)$;
\STATE ~~~~~~~~\textbf{adapt} weights $w^{i*}_j ~ \forall j$;
\STATE ~~~~\textbf{end if}
\STATE ~~~~\textbf{if} $h = \beta h_r$, $\beta = 1, \dots$ \textbf{then}
\STATE ~~~~~~~~\textbf{adapt} model granularity $\rho$;
\STATE ~~~~\textbf{end if}
\STATE \textbf{end for}
\end{algorithmic}
\vspace{3pt}
\hrule
\vspace{4pt}

%\clearpage

\section{Methodology}
\label{sec:meth}

A control-chart-based approach is proposed as a way to label the log stream. Time-windows are classified according to the anomaly severity and standard deviations from the usual system behavior \cite{Decker2}. We provide description of the dataset, attributes, and evaluation measures.

\subsection{Tagging Strategy}

A control chart is a time-series graphic used to monitor a process behavior, phenomenon or variable using the Central Limit Theorem \cite{controlChart}. It is based on the mean $\mu(u)$ of a random variable $u$ that follows a normal distribution \cite{controlChart}. 

Let $\textrm{u}_j = [u_1 ~ \dots ~ u_i ~ \dots ~ u_n]$, $j \geqslant 1$, be a sequence of values that represent the log activity rate, calculated using a fixed $w_j = [\underline{w}_j ~  \overline{w}_j]$; $\textbf{u}_j \in \mathbb{N}^n$. The bounds of the time window, $\underline{w}_j$ and $\overline{w}_j$, are equal to $u_1$ and $u_n$. In addition, let $\mu_j$ be the mean of $\textbf{u}_j$, thus

\vspace{-2pt}

\begin{eqnarray}
\mu_j = \frac{1}{n} \sum\limits_{i=1}^n u_i, ~ u_i \in [\underline{w}_j ~  \overline{w}_j].
\label{eq18}
\end{eqnarray}

\vspace{-2pt}

\noindent Then, $[\mu_1 ~ \dots ~ \mu_j ~ \dots ~ \mu_m]$ is a stream of means.

%\begin{eqnarray}
%\mu = [\mu_1 ~ \dots ~ \mu_j ~ \dots ~ \mu_m].
%\label{eq19}
%\end{eqnarray}

\begin{figure}[htp!]
    \begin{center}
       \includegraphics[width=0.95\columnwidth]{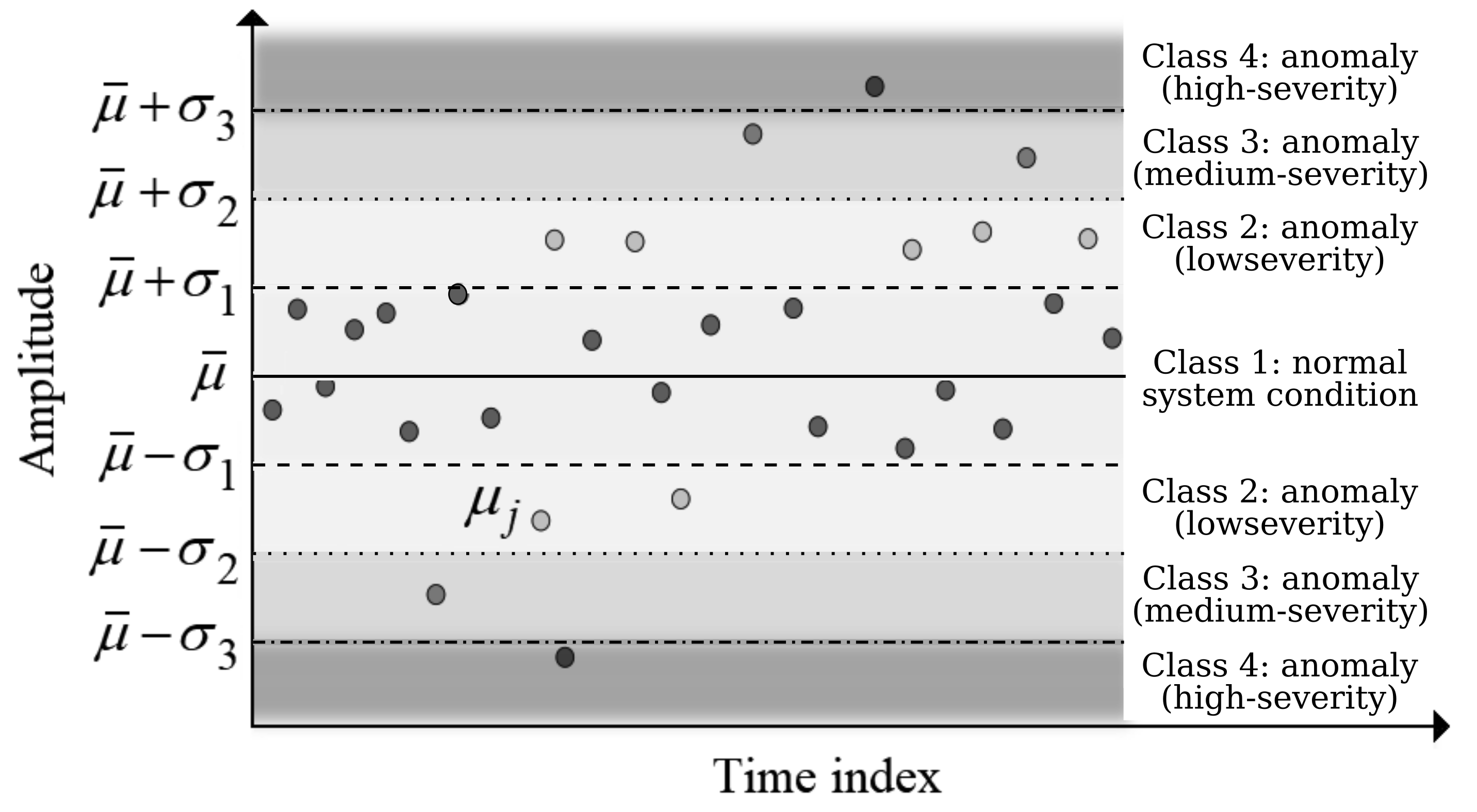}
    \end{center}
	\caption{Control chart procedure to tag log data}
	\label{fig4}
\end{figure}

Figure \ref{fig4} shows how an instance $\mu_j$ is labelled from a control chart. The mean of $\mu$ is

\vspace{-2pt}

\begin{eqnarray}
\overline{\mu} = \frac{1}{m} \sum\limits_{j=1}^{m} \mu_j.
\label{eq20}
\end{eqnarray}
\vspace{-2pt}
\noindent The $k$-th upper and lower horizontal lines in relation to $\overline{\mu}$ refer to the $k$-th standard deviation,

\vspace{-2pt}

\begin{eqnarray}
\sigma_k(\mu) = k* \sqrt{\frac{1}{m}\sum\limits_{j=1}^m (\overline{\mu} - \mu_j)^2}.
\label{eq21}
\end{eqnarray}

\vspace{-2pt}

\noindent If $\mu_j ~ \subset ~ [\overline{\mu} - k ~ \sigma(\mu_ j \forall j), ~ \overline{\mu} + k ~ \sigma(\mu_ j \forall j)]$ and $\mu_j ~ \not\subset ~ [\overline{\mu} - (k-1) ~ \sigma(\mu_ j \forall j), ~ \overline{\mu} + (k-1) ~ \sigma(\mu_ j \forall j)]$, then, for $k = 1$, the instance is tagged as `Class 1', which means normal system operation. If $k = 2, 3, 4$, then $\mu_j$ is tagged as `Class 2', `Class 3' and `Class 4', respectively, see Fig.~\ref{fig4}. These mean low, medium, and high-severity anomaly. The probability that $\mu_j$ is within each class is 67\%, 28\%, 4.7\%, and 0.3\%, respectively. The online anomaly detection problem is unbalanced, and the labels are weak.

%\begin{eqnarray}
%\mu_j ~ \subset ~ [\overline{\mu} - k ~ \sigma(\mu_ %j \forall j), ~ \overline{\mu} + k ~ \sigma(\mu_ j %\forall j)], \label{intclasses}
%\end{eqnarray}

%\noindent for $k = 1$, it is tagged as `Class 1' (normal system condition). Otherwise, if \eqref{intclasses} holds for $k = 2, 3$, and $4$, respectively, $\mu_j$ is tagged as `Class 2', `Class 3', and `Class 4', which mean low, medium, and high-severity anomaly. The greater the value of $k$, the greater the severity of the anomalous behavior.

\vspace{-4pt}

\vspace{-4pt}

% \begin{figure}[htp!]
%     \begin{center}
%       \includegraphics[width=3.6in]{controlChartTagging2.png}
%     \end{center}
% 	\caption{Control chart used to tag mean log data within a time window}
% 	\label{fig4}
% \end{figure}

\subsection{Dataset Description}

StoRM service generates a stream of time-indexed log records. Each log entry is composed by timestamp related to the writing moment, and the message itself. In this work, we are focused on the logging activity, and analysis of the message content is out of the paper scope. 

We extract 5-metrics from the original log file considering fixed time windows, generating vector sequence of 
$\textbf{x} = [x_1 ~~ x_2 ~~ x_3 ~~ x_4 ~~ x_5]$, whose elements are $\overline{\mu}$, $\sigma(\mu_ j \forall j)$, $min(\mu_ j \forall j)$, $max(\mu_ j \forall j)$, and $max(\Delta \mu_j)$. The latter metrics means the maximum difference of the amplitude of two consecutive $\mu_ j$, in which $\mu_ j \subset w_j$.

A vector $\textbf{x}^{[h]}$ is associated to a class label $C = \{1, 2, 3, 4\}$ that, in turn, indicates the system behavior. The true label, $C$, is available after an estimation, $\hat{C}$, is provided by the FBeM or eGNN model. When available, the pair $(\textbf{x},C)^{[h]}$ is used by the learning algorithms for an updating step.

\subsection{Evaluation Measures}

We use three measures to evaluate classifiers performance: (i) classification accuracy, (ii) average number of granules or rules; and (iii) execution time. The former are calculated using recursive relations.

Classification accuracy, $Acc \in [0,1]$, is obtained from
\vspace{-2pt}
\begin{eqnarray}
Acc(new) = \frac{h-1}{h} ~ Acc(old) + \frac{1}{h} ~ \tau,
\end{eqnarray}
\vspace{-2pt}
\noindent in which $\tau = 1$ if $\hat{C}^{[h]} = C^{[h]}$ (right estimation); $\hat{C}^{[h]}$ and $C^{[h]}$ are the estimate and actual classes. Otherwise, $\tau = 0$ (wrong class estimation).

The average number of granules or rules over time, $c_{avg}$, is a measure of model concision. It is computed as
\vspace{-2pt}
\begin{eqnarray}
c_{avg}(new) = \frac{h-1}{h} ~ c_{avg}(old) + \frac{1}{h} ~ c^{[h]}.
\end{eqnarray}
%vspace{5pt}
\vspace{-2pt}

The execution time using a DELL Latitude 5490 64-bit quad-core (Intel Core i58250U, 8GB RAM) is also given.

%\clearpage

\section{Results}
\label{sec:er}

We compare FBeM and eGNN for anomaly classification in a computing center using attributes extracted from logging activity. Experiments do not assume prior knowledge about the data. Classification models are evolved from scratch based on the data stream.

\subsection{Performance Comparison}

We compare the evolving fuzzy and neuro-fuzzy granular classifiers in terms of accuracy, model compactness, and processing time. In a first experiment, we set $h_r$ within $75$ and $125$, $\eta = 3$, and the granularity $\rho^{[0]}$ around $0.3$ and $0.7$. The $\rho$ level evolves during the learning process. These meta-parameters allowed the generation of FBeM and eGNN models with about 10 to 14 rules -- a reasonably compact structure. 

Table \ref{Tab1} shows the results of FBeM and eGNN. Results are averaged over 5 runs of each method for each window length, and shuffled datasets, as described in Sec. \ref{sec:meth}. In other words, for a single window length, a unique dataset is produced, but its instances are shuffled differently, 5 times. The window lengths in minutes are $w = 5,~15,~30,~60$. Each dataset consists of 1,436 instances, 5 attributes, and a class label, which is generated automatically by the control chart approach for model supervision. Four classes are possible. Classes 1 to 4 mean `normal system operation', `low severity', `medium severity', and `high severity' anomaly, respectively.

\begin{table}[!ht]
	\small \caption{FBeM - eGNN Performance in Multiclass Classification of System Anomalies with 99\% of confidence}
	\vspace{-15pt}
	\begin{center}
		\resizebox{\columnwidth}{!}{
			\begin{tabular}{c|ccc}
			    \hline
				\multicolumn{4}{c}{FBeM}\\
				\hline
				Time Window (min) & $Acc$(\%) & \# Rules & Time (s) \\
				\hline
				60 & $85.64 \pm 3.69$ & $12.63 \pm 3.44$ & $0.18 \pm 0.02$ \\
				30 & $75.58 \pm 5.91$ & $10.64 \pm 1.89$ & $0.19 \pm 0.06$ \\
				15 & $66.99 \pm 3.63$ & $~9.94  \pm 1.50$ & $0.18 \pm 0.02$ \\ 
				5  & $67.27 \pm 4.26$ & $13.88 \pm 1.17$ & $0.19 \pm 0.02$ \\
				\hline
				\multicolumn{4}{c}{eGNN}\\
				\hline
				Time Window (min) & $Acc$(\%) & \# Rules & Time (s) \\
				\hline
				60 & $96.17 \pm 0.78$ & $10.35 \pm 1.32 $ & $0.18 \pm 0.02$ \\
				30 & $90.56 \pm 2.70$ & $13.79 \pm 2.26$ & $0.22 \pm 0.04$ \\
				15 & $86.28 \pm 5.29$ & $13.68 \pm 1.31$ & $0.21 \pm 0.02$ \\ 
				5 & $85.00 \pm 2.39$ & $12.61 \pm 0.64$ & $0.22 \pm 0.02$ \\
				\hline
			\end{tabular}
			\label{Tab1}}
	\end{center}
\end{table}

Table \ref{Tab1} shows that the classification performance of both methods increase for larger time windows regarding similar model compactness, i.e., from 10 to 14 rules. The low-pass filter effect strongly impacts the FBeM accuracy, which improves about $18\%$ from the worst to the best case. The best FBeM scenario ($85.64\% \pm 3.69$) is, however, comparable to the worst eGNN scenario ($85.00 \pm 2.39$). The neuro-fuzzy classifer reached a $96.17 \pm 0.78$ accuracy using 60-minute sliding time windows, and about $10$ rules. Compared to FBeM, eGNN uses a higher number of parameters per local model and, therefore, the decision boundaries among classes are more flexible to nonlinearities. Nevertheless, eGNN requires a greater number of instances to update a larger number of parameters. The CPU time is equivalent in all experiments.

\begin{figure}[htp!]
    \begin{center}
       \includegraphics[width=0.95\columnwidth]{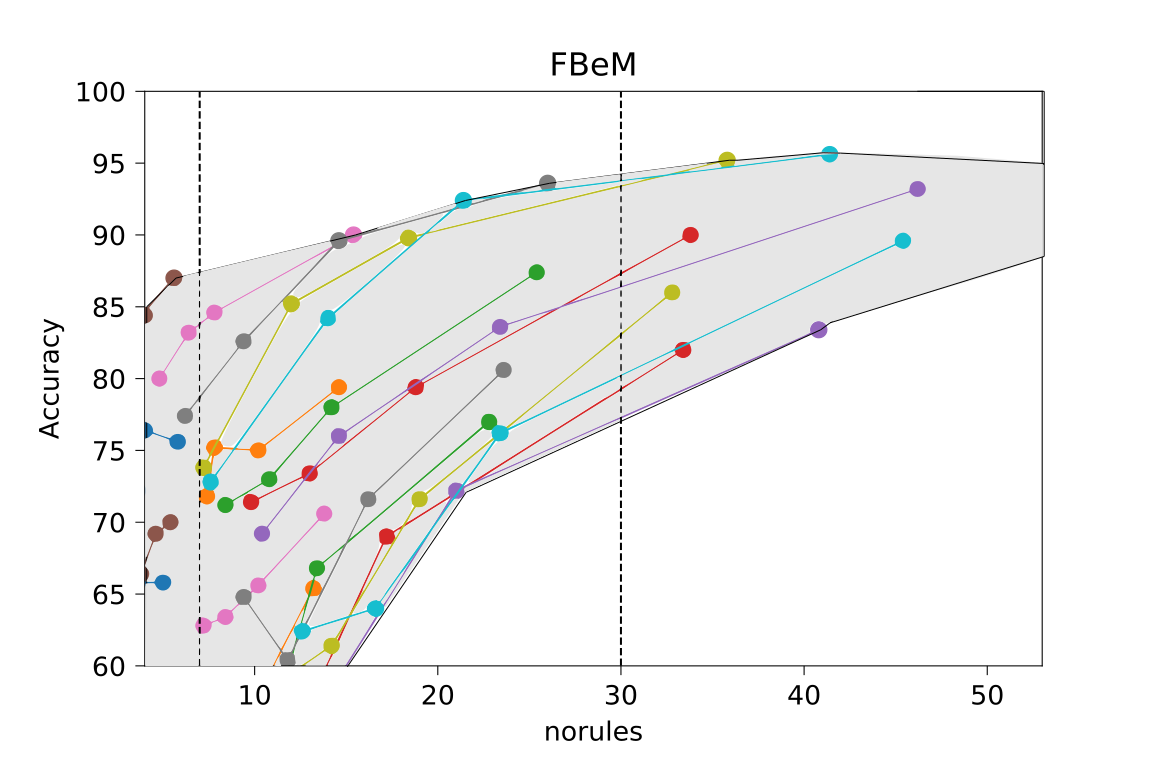}
    \end{center}
	\caption{FBeM Classification accuracy versus amount of rules for an array of meta-parameters}
	\label{rocfbem}
\end{figure}

\begin{figure}[htp!]
    \begin{center}
       \includegraphics[width=0.95\columnwidth]{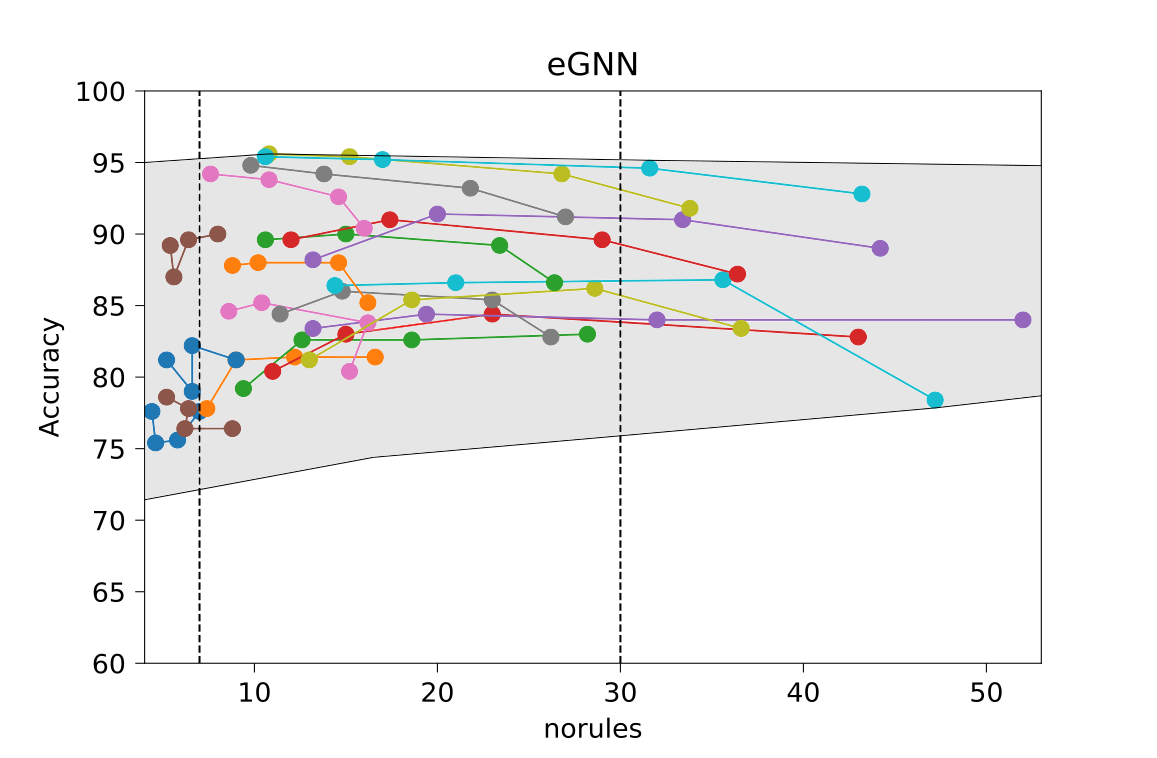}
    \end{center}
	\caption{eGNN classification accuracy versus amount of rules for a variety of meta-parameters}
	\label{rocEGNN}
\end{figure}

A second experiment consists in seeking the best meta-parameters, $\rho$ and $h_r$, of FBeM and eGNN, and, hence, their best performance without upper constraint on model structure. Figures \ref{rocfbem} and \ref{rocEGNN} show the classification accuracy versus number of FBeM and eGNN rules, respectively, for a variety of combinations of meta-parameters. Each point within the gray zone (area of possibility) of the graphs is the average of 5 runs of an algorithm on shuffled data.

\vspace{-8pt}

\vspace{-8pt}

Notice in Fig.~\ref{rocfbem} that FBeM improves its performance if we allow its rule structure to grow. FBeM reaches a $96.1\%$ anomaly classification accuracy using small initial granularities, such as $\rho = 0.1$, and about $40$ rules, whereas the best average performance of eGNN is in fact $96.2\%$ (as shown in the first experiment) using $\rho = 0.7$ and about $10$ rules. Scenarios with few rules may mean that the dynamic of model adaptation is too fast in relation to the dynamic behavior of the target system. In other words, a small amount of granules are dragged to the current scenario, which may be detrimental to the memory of past episodes.

Comparing Figs. \ref{rocfbem} and \ref{rocEGNN}, the relative higher slope of the gray zone of FBeM in relation to eGNN is explained by the higher number of parameters per eGNN local model. Fuzzy neurons and synaptic weights provide eGNN granules with a higher level of local nonlinear ability such that it requires less local models to represent the whole. However, if the eGNN structure grows rapidly, an additional number of parameters is added to the model such that their convergence/maturation to realistic values demand more data instances.

\begin{figure}[htp!]
    \begin{center}
       \includegraphics[width=0.7\columnwidth]{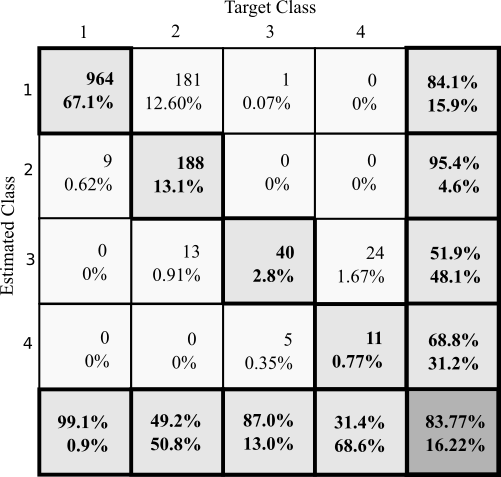}
    \end{center}
	\caption{Example of FBeM confusion matrix in anomaly classification}
	\label{fbemcm}
\end{figure}

\begin{figure}[htp!]
    \begin{center}
       \includegraphics[width=0.72\columnwidth]{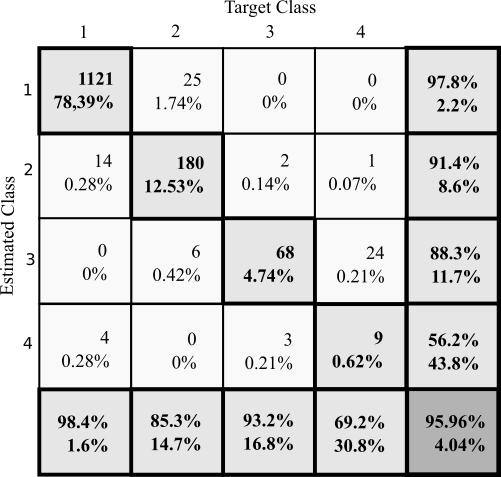}
    \end{center}
	\caption{Example of eGNN confusion matrix in anomaly classification}
	\label{EGNNcm}
\end{figure}

Figures \ref{fbemcm} and \ref{EGNNcm} show an example of confusion matrix for a $83.77\%$-accuracy FBeM and a $95.96\%$-accuracy eGNN. Classes 3 and 4 (medium and high-severity anomalies) are relatively harder to be identified. Notice that confusion in FBeM happens, in general, in the neighbourhood of a target class, whereas often eGNN estimates an instance as being `Class 4' when it is `Class 1'. This indicates that a local model may still be learning values for itself. FBeM slightly outperformed eGNN regarding detection of the most critical `high severity' class. eGNN's overall accuracy is superior.

\vspace{-5pt}

\section{Conclusion}
\label{sec:cfw}

We proposed and compared evolving fuzzy and neuro-fuzzy granular classifiers, FBeM and eGNN, in a real-time anomaly detection problem considering online log data streams from the main Italian computing center that supports CERN, the WLCG Tier-1 at Bologna. Logging activity rate is a standard behavioral metric obtained from online log data.

The accuracy of both, FBeM and eGNN, presented an increasing trend with increasing lengths of sliding time windows. Larger time windows work as low-pass filters of higher order; they suppress stochastic components. In a first experiment, the best average FBeM accuracy, $85.6\%$, has shown to be comparable to the worst average eGNN accuracy, $85.0\%$, with $99\%$ confidence, using similar model structures with about 12 rules. The best eGNN result uses 60-minute time windows and offer $96.2\%$ classification accuracy. 

A second experiment, which seeks the best initial meta-parameters of FBeM and eGNN, shows that, when we do not establish a rigorous upper bound for the average number of rules, FBeM reaches a $96.1\%$ anomaly classification accuracy using small initial granularities, such as $\rho = 0.1$, and about $40$ rules, whereas the best average performance of eGNN is in fact $96.2\%$ using $\rho = 0.7$ and about $10$ rules. We conclude that the use of fuzzy neurons and synaptic weights provides eGNN granules with a higher level of local nonlinear capacity such that the neuro-fuzzy model requires less local elements to represent the data. Interestingly, FBeM slightly outperformed eGNN regarding detection of the most critical class, which corresponds to `high severity' anomalous windows.

In general, both classifiers are suitable to be applied to the WLCG Italian Tier-1 to monitor the services that produce event-oriented log data. Evolving granular classifiers provide a real-time log-content processing approach, which is the key to assist decision making toward predictive maintenance.

\bibliographystyle{ieeetr}

\bibliography{biblio}

\begin{thebibliography}{10}

\bibitem{Skrjanc1}
I.~Škrjanc, J.~Iglesias, A.~Sanchis, D.~Leite, E.~Lughofer, and F.~Gomide,
  ``Evolving fuzzy and neuro-fuzzy approaches in clustering, regression,
  identification, and classification: A survey,'' {\em Inf. Sci.}, vol.~490,
  pp.~344--368, 2019.

\bibitem{Cordovil}
L.~A. Cordovil, P.~H. Coutinho, I.~Bessa, M.~F. D'Angelo, and R.~Palhares,
  ``Uncertain data modeling based on evolving ellipsoidal fuzzy information
  granules,'' {\em IEEE Transactions on Fuzzy Systems}, p.~11p. DOI:
  doi.org/10.1109/TFUZZ.2019.2937052, 2019.

\bibitem{Garcia}
C.~Garcia, D.~Leite, and I.~Škrjanc, ``Incremental missing-data imputation for
  evolving fuzzy granular prediction,'' {\em IEEE T Fuzzy Syst.}, pp.~1--15,
  2019.
\newblock DOI: 10.1109/TFUZZ.2019.2935688.

\bibitem{Hyde}
R.~Hyde, P.~Angelov, and A.~Mackenzie, ``Fully online clustering of evolving
  data streams into arbitrarily shaped clusters,'' {\em Inf. Sci.}, vol.~382,
  pp.~1--41, 2016.

\bibitem{SilvaP}
P.~Silva, H.~Sadaei, R.~Ballini, and F.~Guimaraes, ``Probabilistic forecasting
  with fuzzy time series,'' {\em IEEE Transactions on Fuzzy Systems}, p.~14p.
  DOI: doi.org/10.1109/TFUZZ.2019.2922152, 2019.

\bibitem{Venkatesan}
R.~Venkatesan, M.~Er, M.~Dave, M.~Pratama, and S.~Wu, ``A novel online
  multi-label classifier for high-speed streaming data applications,'' {\em
  Evolving Systems}, pp.~303--315, 2016.

\bibitem{Souza}
P.~V. Souza, T.~Rezende, A.~Guimaraes, V.~Araujo, L.~Batista, G.~Silva, and
  V.~Silva, ``Evolving fuzzy neural networks to aid in the construction of
  systems specialists in cyber attacks,'' {\em Journal of Intelligent \& Fuzzy
  Systems}, vol.~36, no.~6, pp.~6773--6763, 2019.

\bibitem{Bezerra}
C.~Bezerra, B.~Costa, L.~A. Guedes, and P.~Angelov, ``An evolving approach to
  data streams clustering based on typicality and eccentricity data
  analytics,'' {\em Information Sciences}, vol.~518, pp.~13--28, 2020.

\bibitem{Edw2020}
M.~Pratama, W.~Pedrycz, and E.~Lughofer, ``Online tool condition monitoring
  based on parsimonious ensemble+,'' {\em IEEE T Cybernetics}, vol.~50, no.~2,
  pp.~664--677, 2020.

\bibitem{chep}
A.~Di~Girolamo~et al., ``Operational intelligence for distributed computing
  systems for exascale science,'' in {\em 24th Int. Conf. on Computing in High
  Energy and Nuclear Physics (CHEP), AU}, pp.~1--8, 2020.

\bibitem{Leite10}
D.~Leite, R.~Ballini, P.~Costa~Jr, and F.~Gomide, ``Evolving fuzzy granular
  modeling from nonstationary fuzzy data streams,'' {\em Evolving Systems},
  vol.~3, pp.~65--79, 2012.

\bibitem{Leite14}
D.~Leite, P.~Costa~Jr, and F.~Gomide, ``Evolving granular neural networks from
  fuzzy data streams,'' {\em Neural Networks}, vol.~38, pp.~1--16, 2013.

\bibitem{Giommi}
L.~Giommi~et al., ``Towards predictive maintenance with machine learning at the
  {INFN-CNAF} computing centre,'' in {\em Int. Symp. on Grids \& Clouds (ISGC).
  Taipei: Proceedings of Science}, pp.~1--17, 2019.

\bibitem{Diotalevi}
T.~Diotalevi {\em et~al.}, ``Collection and harmonization of system logs and
  prototypal analytics services with the elastic (elk) suite at the infn-cnaf
  computing centre,'' in {\em Int. Symp. on Grids \& Clouds (ISGC). Taipei:
  Proceedings of Science}, pp.~1--15, 2019.

\bibitem{Rossi}
S.~R. Tisbeni, ``Big data analytics towards predictive maintenance at the
  {INFN-CNAF} computing centre,'' Master's thesis, U. of Bologna, 2019.

\bibitem{FUZZlog}
L.~Decker, D.~Leite, L.~Giommi, and D.~Bonacorsi, ``Real-time anomaly detection
  in data centers for log-based predictive maintenance using an evolving
  fuzzy-rule-based approach,'' in {\em IEEE World Congress on Computational
  Intelligence (WCCI, FUZZ-IEEE), Glasgow}, p.~8p, 2020.

\bibitem{Minarini}
F.~Minarini, ``Anomaly detection prototype for log-based predictive maintenance
  at {INFN-CNAF},'' Master's thesis, U. of Bologna, 2019.

\bibitem{Soares2}
E.~Soares, P.~Costa, B.~Costa, and D.~Leite, ``Ensemble of evolving data clouds
  and fuzzy models for weather time series prediction,'' {\em Appl. Soft
  Comput.}, vol.~64, pp.~445--453, 2018.

\bibitem{Shaker}
A.~Shaker and E.~Lughofer, ``Self-adaptive and local strategies for a smooth
  treatment of drifts in data streams,'' {\em Evolving Systems}, vol.~5, no.~4,
  pp.~239--257, 2014.

\bibitem{Gu}
X.~Gu and P.~P. Angelov, ``Self-organising fuzzy logic classifier,'' {\em Inf.
  Sci.}, vol.~447, pp.~36 -- 51, 2018.

\bibitem{Mirza}
B.~Mirza, Z.~Lin, and N.~Liu, ``Ensemble of subset online sequential extreme
  learning machine for class imbalance and concept drift,'' {\em
  Neurocomputing}, vol.~149, pp.~316 -- 329, 2015.
\newblock Advances in neural networks Advances in Extreme Learning Machines.

\bibitem{MPratama}
M.~Pratama, J.~Lu, E.~Lughofer, G.~Zhang, and S.~Anavatti, ``Scaffolding type-2
  classifier for incremental learning under concept drifts,'' {\em
  Neurocomputing}, vol.~191, pp.~304 -- 329, 2016.

\bibitem{Charles}
C.~Aguiar and D.~Leite, ``{Unsupervised Fuzzy eIX}: Evolving internal-external
  fuzzy clustering,'' in {\em IEEE Evolving and Adaptive Intelligent Systems
  (EAIS 2020), Bari - IT}, pp.~1--8, 2020.

\bibitem{Kim}
Y.~Kim and C.~Park, ``An efficient concept drift detection method for streaming
  data under limited labeling,'' {\em IEICE Transactions on Information and
  Systems}, vol.~E100.D, pp.~2537--2546, 2017.

\bibitem{LeiteGra}
D.~Leite, P.~Costa, and F.~Gomide, ``Granular approach for evolving system
  modeling,'' in {\em Computational Intelligence for Knowledge-Based Systems
  Design} (E.~H{\"u}llermeier, R.~Kruse, and F.~Hoffmann, eds.), (Berlin,
  Heidelberg), pp.~340--349, Springer, 2010.

\bibitem{eogs}
D.~Leite, G.~Andonovski, I.~Skrjanc, and F.~Gomide, ``Optimal rule-based
  granular systems from data streams,'' {\em IEEE Transactions on Fuzzy
  Systems}, vol.~28, no.~3, pp.~583--596, 2020.

\bibitem{GranularHandBook}
W.~Pedrycz, A.~Skowron, and V.~Kreinovich, {\em Handbook of Granular
  Computing}.
\newblock USA: Wiley-Interscience, 2008.

\bibitem{Pedrycz}
W.~Pedrycz and F.~Gomide, {\em {An Introduction to Fuzzy Sets: Analysis and
  Design}}.
\newblock MIT Press, 2000.

\bibitem{Beliakov}
G.~Beliakov, A.~Pradera, and T.~Calvo, ``Aggregation functions: A guide for
  practitioners,'' in {\em Studies in Fuzziness and Soft Computing}, 2007.

\bibitem{Decker2}
L.~D. {de Sousa}~et al., ``Event detection framework for wireless sensor
  networks considering data anomaly,'' in {\em 2012 IEEE Symposium on Computers
  and Communications (ISCC)}, pp.~000500--000507, July 2012.

\bibitem{controlChart}
P.~Qiu, {\em Introduction to Statistical Process Control}.
\newblock Wiley: India, 2014.

\end{thebibliography}

\end{document}